\documentclass[a4paper, 10pt, conference]{ieeeconf}      

\IEEEoverridecommandlockouts                              

\usepackage[dvipdfmx]{graphicx} 

\title{\LARGE \bf
Team Irisapu Project Description for DRC2023
}

\author{Reon Ohashi$^{1}$, Shinjitsu Agatsuma$^{1}$, Kazuya Tsubokura$^{2}$ and Yurie Iribe$^{1}$
\thanks{$^{1}$ School of Information Science and Technology, 
	Aichi Prefectural University, 1522-3 Ibaragabasama, Nagakute-shi, Aichi, 480-1198, Japan
	{\tt\small \{is211002, is211014\}@cis.aichi-pu.ac.jp, iribe@ist.aichi-pu.ac.jp}}
\thanks{$^{2}$ Graduate School of Information Science and Technology, 
	Aichi Prefectural University, 1522-3 Ibaragabasama, Nagakute-shi, Aichi, 480-1198, Japan
	{\tt\small id231001@cis.aichi-pu.ac.jp}}
}

\begin{document}

\maketitle
\thispagestyle{empty}
\pagestyle{empty}

\begin{abstract}

This paper describes the dialog robot system designed by Team Irisapu for the preliminary round of the Dialogue Robot Competition 2023 (DRC2023).
In order to generate dialogue responses flexibly while adhering to predetermined scenarios, we attempted to generate dialogue response sentences using OpenAI's GPT-3. 
We aimed to create a system that can appropriately respond to users by dividing the dialogue scenario into five sub-scenarios, and creating prompts for each sub-scenario.
Also, we incorporated a recovery strategy that can handle dialogue breakdowns flexibly. 
Our research group has been working on research related to dialogue breakdown detection, and we incorporated our findings to date in this competition.
As a result of the preliminary round, a bug in our system affected the outcome and we were not able to achieve a satisfactory result. 
However, in the evaluation category of "reliability of provided information", we ranked third among all teams.
\end{abstract}

\section{Introduction}

This paper describes the travel agent dialog robot system designed by Team Irisapu for the preliminary round of the Dialogue Robot Competition 2023 (DRC2023),  an competition evaluating the performance of dialogue robots \cite{drc2023}. 
We are also participating in DRC2022 \cite{drc2022, drc}, but we significantly changed our design policy for this competition. In DRC2022, we reduced dialogue breakdowns and discontinuity by having sentences created manually in advance read aloud \cite{irisapu2022}. However, while DRC2022 was a task to recommend one of two tourist spots, DRC2023 was updated to a task of devising a travel plan around Kyoto city, so the cost of creating sentences to be read aloud in advance was enormous and could not be covered completely. With the recent rise of LLM, it has become possible to generate very natural sentences, so in this competition, we used OpenAI's GPT-3 to try to generate dialogue response sentences. In addition, we focused on dialogue breakdown detection, which we are working on in our lab, and we also performed dialogue breakdown detection during dialogue.

The flow of the dialogue scenario is explained in Section \ref{sec2}. 
The dialogue breakdown detection will be explained in section \ref{sec3}.
The results of the evaluations provided by users are reported and discussed in Section \ref{sec4}, and the findings of this paper are summarized in Section \ref{sec5}.

\begin{figure}[thpb]
  \centering
  \includegraphics[scale=0.04]{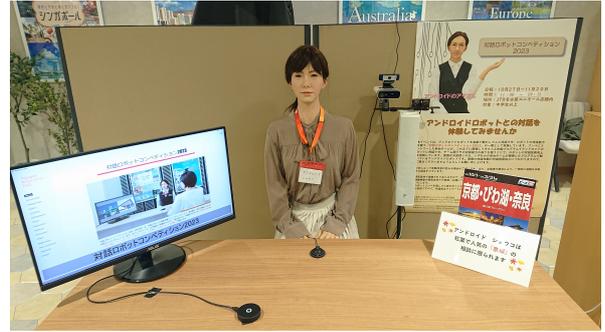}
  \caption{Dialogue Robot}
  \label{fig:robot}
\end{figure}

\section{Dialogue Flow} \label{sec2}

\begin{figure*}[thpb]
  \centering
  \includegraphics[scale=0.4]{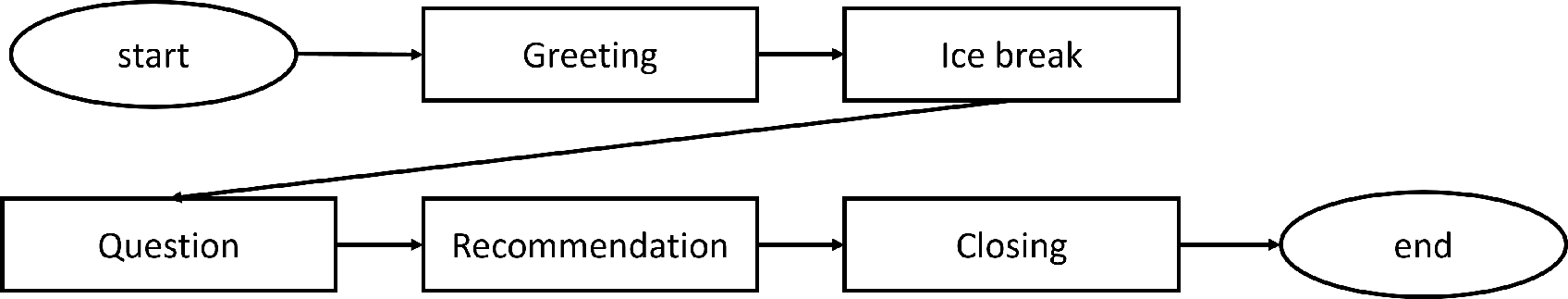}
  \caption{Dialogue flow}
  \label{fig:dialog_flow}
\end{figure*}

This chapter explains the flow of the dialogue scenario we constructed for DRC2023. The flow of the dialogue is shown in figure \ref{fig:dialog_flow}. The scenario is composed of five main elements. We attempted to generate utterances that followed the scenario by changing the prompt description in each phase. First, the robot initiates a greeting in the greeting phase, and then in the ice break phase, it makes small talk to reduce the psychological distance with the customer. Then, in the question phase, the robot asks the customer several questions about their trip. Based on the information obtained from these questions, the robot makes recommendations about tourist spots in the recommendation phase. Finally, the dialogue is ended in the closing phase.

\subsection{Greeting}

The greeting phase is the first conversation the robot has with the customer. In this phase, we designed a prompt with the intention of "generating an utterance to greet the customer and ask for their name".

\subsection{Ice break}

In the ice break phase, the robot makes small talk about travel with the customer to foster a closer relationship. In this phase, we designed a prompt to initiate small talk about Kyoto, the travel destination.

\subsection{Question}

In the question phase, the robot inquires the customer for information necessary to recommend a sightseeing plan. In this phase, the customer is asked about three question items: the purpose of their travel, who they are travelling with, and the season they want to travel in.

\subsection{Recommendation}

In the recommendation phase, the robot explains the recommended sightseeing spots to the customer. In this phase, we designed a prompt to select recommended sightseeing locations from the conversation so far, and introduce those locations.

\subsection{Closing}

In the closing phase, utterances are made to finish the conversation.

\section{Dialogue Breakdown Detection and Recovery} \label{sec3}

Even with the use of GPT-3, dialogue may still break down due to factors such as speech recognition errors. Hence, we decided to perform dialogue breakdown detection at every turn and determine whether or not a breakdown has occurred. Various methods have been proposed for dialogue breakdown detection \cite{sugiyama2021,tsubokura2022}, but for simplicity, we decided to judge the presence or absence of a breakdown using GPT-3's prompt. Although our previous study about breakdown detection \cite{tsubokura2022} uses multi-modal user data for breakdown detection and utilizing multimodal information is important for accurate breakdown determination, in this task we only use text due to the need for real-time determination during the dialogue. If a breakdown is identified, we decided to have the robot utter the following set phrase: ``I'm sorry. Did I say something wrong? Or I may not have been able to catch our conversation. Could you please tell me again?''.

\section{Result of the preliminary round} \label{sec4}

In DRC2023, each team's system is evaluated based on a combination of two factors: an impression evaluation based on nine questionnaire items, and an evaluation of the proposed travel plan. For detailed information on the evaluation items, please refer to \cite{drc2023}. Our team was able to conduct evaluations with 15 participants during the preliminary round period. The results of the impression evaluation are shown in table \ref{tab:result}. Avg. represents the average of evaluation scores and rank represents our team's ranking among 13 teams. Regarding the evaluation of the travel plan, we ranked 12th among the 13 teams.

Overall, our system resulted in a very poor evaluation. The cause for this is clear; there was a fatal bug in our system. Due to this system bug, it took more than one minute for our system to respond to the experimenter's speech recognition results. As a result, our system received low ratings on many items including naturalness and reliability of robot. The bug also prevented the system from sufficiently explaining the plan within the allocated time, resulting in a low evaluation for the plan.

On the other hand, we ranked third among all teams in terms of the reliability of the information provided. We are also surprised by this result, but it is possible that our two strategies, natural response sentences generated by GPT-3 and recovery by dialogue breakdown detection, brought about some effect on this indicator.

\begin{table}[h]
\caption{Results of the Questionnaire}
\label{tab:result}
\begin{center}
\begin{tabular}{ccc} \hline
Item & Avg. & rank \\\hline \hline
Information & 4.27 & 8 \\ \hline
Naturalness & 2.67 & 12 \\ \hline
Appropriateness & 4.07 & 9 \\ \hline
Likeability & 4.40 & 8 \\ \hline
Satisfaction with dialogue & 3.93 & 7 \\ \hline
Reliability of robot & 3.80 & 10 \\ \hline
Reference level of information & 4.00 & 10 \\ \hline
Reliability of information & 5.60 & 3 \\ \hline
How much do you want to come back? & 3.80 & 9 \\ \hline
\end{tabular}
\end{center}
\end{table}

\section{Conclusions} \label{sec5}

In this paper, we discussed the design of team irisapu's system, which competed in DRC2023. In the preliminaries, we were not able to obtain the evaluation results we expected due to a bug in our system. However, considering that our system received a high evaluation in terms of the reliability of information, we believe we have proposed a travel agency robot dialogue system with certain capabilities.


\end{document}